\documentclass[10pt,twocolumn]{article}

\usepackage[T1]{fontenc}
\usepackage[utf8]{inputenc}
\usepackage[english]{babel}
\usepackage{lmodern}
\usepackage{microtype}
\usepackage[margin=0.72in,columnsep=0.25in]{geometry}
\usepackage{graphicx}
\usepackage{booktabs}
\usepackage{tabularx}
\usepackage{enumitem}
\usepackage{amsmath}
\usepackage{amssymb}
\usepackage{xcolor}
\usepackage{url}
\usepackage{natbib}
\usepackage{balance}
\usepackage[hidelinks]{hyperref}

\definecolor{sectionblue}{HTML}{183B61}
\definecolor{rulegray}{HTML}{B7C4D3}
\definecolor{softblue}{HTML}{EAF2FC}

\hypersetup{
  pdftitle={From Conceptual EneMonIE to an Executable Backend},
  pdfauthor={Janik Lesage, Khadim L\^o, Jean-S\'ebastien Dessureault},
  pdfsubject={Energy information extraction architecture},
  pdfkeywords={energy informatics, information extraction, semantic pipeline, question answering, software architecture}
}

\setlist[itemize]{leftmargin=*,topsep=3pt,itemsep=2pt,parsep=0pt}
\setlist[enumerate]{leftmargin=*,topsep=3pt,itemsep=2pt,parsep=0pt}

\makeatletter
\renewcommand{\section}{\@startsection{section}{1}{\z@}%
  {-2.3ex plus -0.8ex minus -.2ex}%
  {1.0ex plus .2ex}%
  {\normalfont\large\bfseries\color{sectionblue}}}

\renewcommand{\maketitle}{%
  \twocolumn[{%
    \begin{center}
      {\LARGE\bfseries From Conceptual EneMonIE to an Executable Backend:\\[3pt]
      A Modular Architecture for Energy Information Extraction and Grounded Question Answering\par}
      \vspace{0.8em}
      {\large Janik Lesage, Khadim L\^o, and Jean-S\'ebastien Dessureault\par}
      \vspace{0.25em}
      {\normalsize D\'epartement de math\'ematiques et d'informatique,\\
      Universit\'e du Qu\'ebec \`a Trois-Rivi\`eres, Canada\par}
      \vspace{0.7em}
      \rule{0.96\textwidth}{0.5pt}
    \end{center}
    \vspace{0.2em}
  }]%
}
\makeatother

\newcommand{\system}{EneMonIE-Backend}
\newcommand{\code}[1]{\texttt{#1}}

\begin{document}

\maketitle

\begin{abstract}
The original EneMonIE proposal describes a Web-based semantic system for continuously monitoring energy trends through targeted information extraction from heterogeneous online media. This article presents \system, an implementation-derived redesign that converts that conceptual pipeline into a modular, executable, and application-independent backend. The architecture separates five concerns: data sources, acquisition and normalization, semantic processing, knowledge organization, and persistent storage. It operationalizes configurable connectors for RSS feeds, URL collections, arXiv, social-media JSON files, and open-data CSV files; it also introduces question-driven Web search and scraping. Relevant documents pass through energy categorization, named-entity recognition, temporal and event extraction, source-sensitive sentiment analysis, and extractive summarization. The resulting semantic representations support entity co-occurrence graphs, cross-document interlinking, cluster-level information fusion, and relational persistence in SQLite. Three grounded question-answering paths are exposed at the backend level: live Web evidence, the stored corpus, and selected configured resources. Answer generation uses a language model when configured and otherwise falls back to deterministic extractive output. Unlike the original proposal, the presented architecture deliberately excludes the user-interface and server-side application layers, allowing the semantic core to be embedded in different applications. A deterministic smoke validation confirms relevance filtering, persistence, clustering, evidence retrieval, and document deletion. The work provides a reproducible architectural basis for future evaluation on larger multilingual energy corpora.
\end{abstract}


\section{Introduction}

Energy systems generate a continuous stream of heterogeneous information: news articles, research papers, public datasets, policy announcements, organizational reports, and social-media posts. The value of these materials lies not only in their individual content but also in their relationships. A power-grid investment reported by a utility, discussed by the public, and evaluated in a research preprint may represent a single underlying development. Manually monitoring such evidence is costly, slow, and difficult to reproduce.

The conceptual EneMonIE architecture was originally proposed by Dilek Küçük \citep{kucuk2022}. The present work implements, restructures, and extends that proposal as an application-independent backend. The original conceptual design combines text categorization, named-entity recognition (NER), temporal-expression and event extraction, sentiment analysis, social-network construction, media interlinking, information fusion, summarization, a central database, and Web-based retrieval and visualization. It also emphasizes modularity, bilingual processing, and the reuse of domain ontologies and lexical resources.

The original proposal is deliberately broad. It specifies the semantic functions that the system should provide, but it does not prescribe an executable control flow, concrete storage schema, connector contract, question-answering strategy, or graceful degradation mechanism. These implementation decisions become essential when the architecture is turned into software. They determine how duplicate documents are handled, how relevance is enforced before expensive analysis, how semantic objects are represented, and how evidence is exposed to downstream consumers.

This article presents \system, a backend architecture derived from a working Python implementation of the EneMonIE\cite{kucuk2022} concept. The central design decision is to isolate the semantic and data-management core from any particular application. FastAPI endpoints, a desktop client, an HTML dashboard, and Web visualization are therefore outside the boundary of the architecture described here. The backend can instead be invoked by a command-line process, a scheduler, a test harness, or a future service layer.

The main contributions are as follows: 1) an executable five-layer decomposition of the original EneMonIE concept. 2) normalized ingestion through configurable source adapters and stable document identifiers. 3) an explicit semantic-processing sequence with an early energy-relevance gate. 4) operational definitions for entity graphs, media similarity, clustering, and fusion. 5) a normalized SQLite schema connecting documents, semantic annotations, entity edges, and clusters. 6) three evidence-grounded question-answering paths with a local fallback when no external language model is configured. 7) an application-independent boundary that separates reusable domain logic from presentation and transport concerns.

\begin{figure*}[t]
  \centering
  \includegraphics[width=0.97\textwidth]{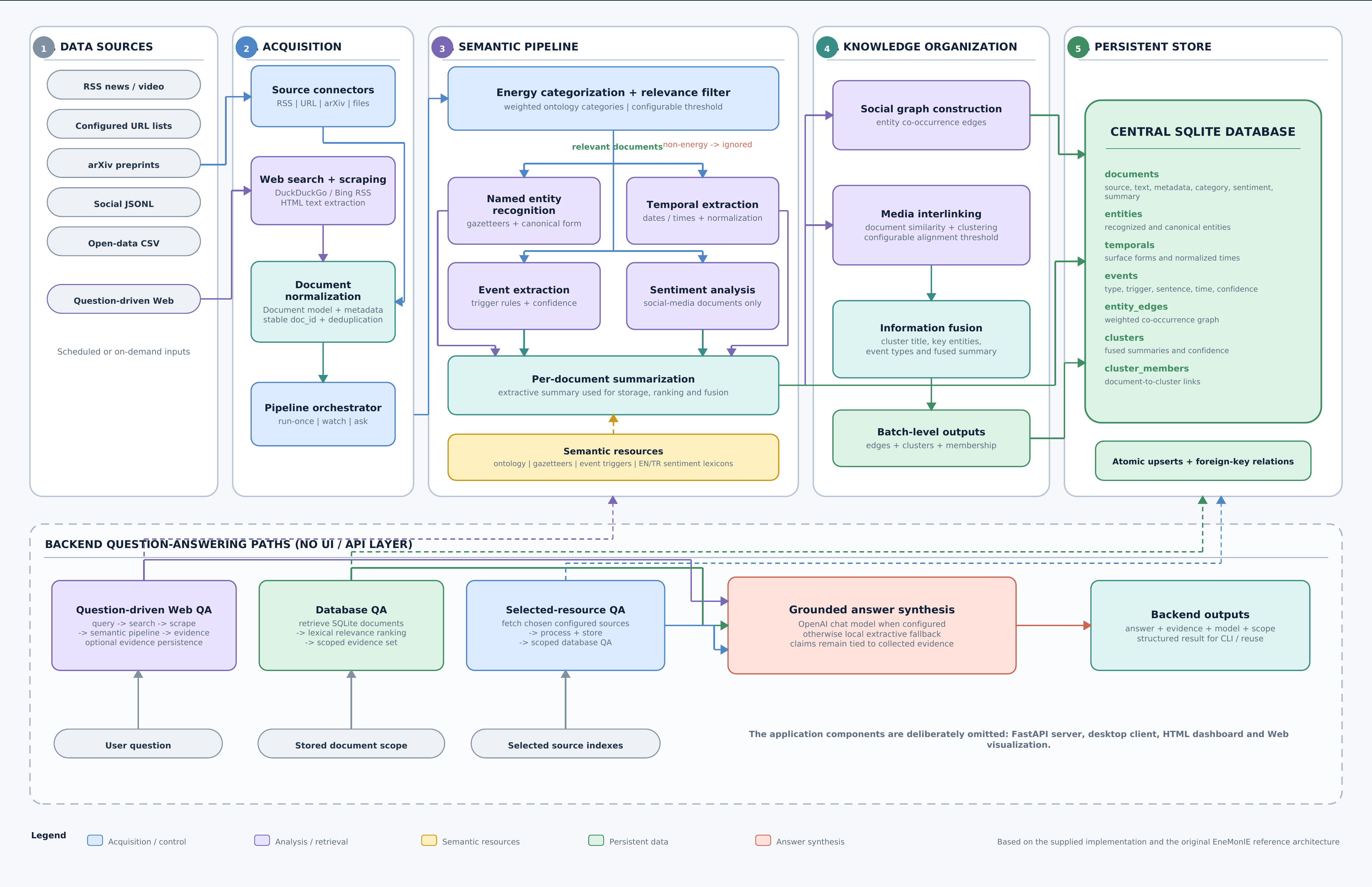}
  \caption{Implementation-derived architecture of \system. The architecture is conceptually adapted and substantially extended from the original EneMonIE proposal by \citet{kucuk2022}.}
  \label{fig:architecture}
\end{figure*}

\section{Background and Design Motivation}

\subsection{From semantic monitoring to an executable pipeline}

EneMonIE belongs to the broader family of continuous knowledge-acquisition systems. Never-Ending Language Learning (NELL) emphasized persistent learning from Web content \citep{carlson2010}; DeepDive demonstrated large-scale knowledge-base construction using statistical learning and inference \citep{niu2012}; and Knowledge Vault combined Web extraction with probabilistic knowledge fusion \citep{dong2014}. EneMonIE narrows this general objective to the energy domain, where entities, events, dates, technologies, organizations, and public opinions must be related across sources.

Several components of the original design are well-established NLP tasks. NER over informal text is challenging because social media frequently departs from conventional spelling and syntax \citep{ritter2011}. Temporal normalization links textual expressions to comparable dates \citep{stroetgen2010}, while event extraction structures occurrences that may have policy, operational, environmental, or technological consequences \citep{ritter2012}. Sentiment analysis provides a coarse representation of expressed opinion \citep{pang2008}, and multi-document or social-media summarization reduces repeated evidence to concise statements \citep{chakrabarti2011}. Energy ontologies further provide domain vocabulary and conceptual structure \citep{kucuk2015}.

An executable architecture must coordinate these tasks and define their data contracts. In particular, downstream clustering should not operate on raw text alone when normalized entities, event types, and dates are available. Similarly, question answering should not bypass the semantic pipeline when live Web pages are collected. \system{} addresses these concerns by using one normalized document model and one processing function for both scheduled ingestion and question-driven evidence collection.

\subsection{Design requirements}

The redesign follows six requirements: 1) \textbf{Modularity.} Source-specific acquisition and semantic analyzers must be replaceable behind stable interfaces. 2) \textbf{Early filtering.} Non-energy documents must be rejected before semantic extraction and persistence. 3) \textbf{Traceability.} Every derived object must remain connected to its source document and, where available, its original URL. 4) \textbf{Determinism by default.} Core extraction, storage, ranking, and fallback answering must operate without a remote language model. 5) \textbf{Evidence grounding.} Generated answers must be based on a bounded evidence set rather than unconstrained model knowledge, following the grounding motivation of retrieval-augmented generation \citep{lewis2020}. 6) \textbf{Application independence.} Transport and presentation layers must not own the semantic pipeline or the database model.

\section{Proposed Architecture}

Figure~\ref{fig:architecture} presents the complete backend architecture. The main pipeline consists of five layers, while a separate lower section illustrates question-driven execution paths that reuse the same processing and persistence services. More specifically, the upper pipeline is responsible for acquiring, normalizing, analyzing, organizing, and storing energy-related documents, whereas the lower section presents three backend question-answering workflows. User-interface, web-server, and visualization components are intentionally excluded from the architectural design.

\subsection{Layer 1: heterogeneous data sources}

The first layer represents both configured and on-demand inputs. Configured sources include RSS news or video feeds, lists of Web pages, arXiv queries, social-media records stored as JSON Lines, and open energy data represented as CSV. These inputs cover the principal media types identified in the original EneMonIE proposal while making each acquisition mechanism explicit.

Question-driven Web retrieval is modeled separately because its lifecycle differs from scheduled polling. Instead of collecting all items from a predeclared source, it transforms a question into an energy-oriented query, obtains candidate results, downloads a bounded number of pages, and submits those pages to the standard semantic pipeline. This design prevents the question-answering path from becoming an unfiltered shortcut around EneMonIE processing.

\subsection{Layer 2: acquisition and normalization}

All configured connectors implement a common fetch operation and return a normalized \code{Document}. A document contains a stable identifier, source type, source name, title, text, URL, language, publication time, and arbitrary metadata. Source adapters hide format-specific details such as RSS XML, arXiv Atom records, HTML cleaning, JSONL decoding, and CSV row serialization.

Normalization is followed by identifier-based deduplication. Within a collection run, documents are indexed by \code{doc\_id}; repeated identifiers overwrite earlier occurrences before semantic processing. At storage time, the same identifier becomes the relational primary key and enables idempotent upserts. This two-stage strategy reduces duplicate processing while supporting updated representations of a previously observed source.

The orchestrator exposes three operational modes. A one-shot mode executes a complete collection and processing cycle; a watch mode repeats the cycle at a configurable interval; and question-oriented modes collect or retrieve only the evidence required for a request. Configuration controls the database path, relevance and alignment thresholds, source definitions, model selection, and Web-retrieval limits.

\subsection{Layer 3: semantic processing}

\subsubsection{Energy relevance gate}

The categorizer tokenizes each document and scores ontology categories with weighted keyword counts. For category $c$ and document $d$, the implemented score is

\begin{equation}
S(c,d)=\sum_{k\in K_c} w_{c,k}\,n(k,d),
\label{eq:category}
\end{equation}

where $K_c$ is the category keyword set, $w_{c,k}$ is the ontology weight, and $n(k,d)$ is the number of exact token occurrences. The highest-scoring category is selected. A document is rejected if no category matches or if its maximum score falls below the configured relevance threshold. Current categories cover renewable energy, grid operations, policy and markets, conventional energy, and research and technology.

Placing this gate before the extractors has two benefits. First, it makes the stored corpus domain-specific. Second, it avoids running later components on evidently unrelated material. The threshold is externalized because the appropriate precision--recall trade-off depends on the deployment corpus.

\subsubsection{Parallel semantic annotations}

Relevant documents are passed to four analyzers: 1) \textbf{NER} combines energy gazetteers with rule-based patterns. Entities are assigned labels such as energy organization, location, technology, company, feeder, power plant, or miscellaneous proper name. Canonical lowercase forms support comparison across documents. 2) \textbf{Temporal extraction} detects supported date formats and weekdays, then normalizes parseable values to ISO timestamps. 3) \textbf{Event extraction} uses domain trigger lexicons for outages, plant commissioning, policy changes, investments, accidents, and research breakthroughs. Confidence is increased when a sentence includes reporting cues, a normalized time, or sufficient context. 4) \textbf{Sentiment analysis} is applied only to social-media documents. English and French lexicons produce a positive, negative, or neutral label from the difference between positive and negative token counts.

These analyzers are logically parallel: each receives the normalized document, although event extraction also consumes the temporal output to associate a normalized time. Their outputs are combined into a \code{ProcessedDocument}. A frequency-based extractive summarizer selects a small number of high-scoring sentences while preserving their original order.

Semantic resources are stored outside the processing classes as YAML files. The current resources include the energy ontology, entity gazetteers, event triggers, and English and French sentiment lexicons. This separation enables vocabulary updates without changing orchestration code.

\subsection{Layer 4: knowledge organization}

The knowledge-organization layer operates across processed documents. First, the social-graph builder creates one undirected edge for each pair of unique canonical entities that co-occurs in a document. Edge weight is the number of documents containing the pair. The result is a compact association graph that can be analyzed or visualized by downstream software.

Second, the media interlinker estimates whether two documents discuss the same development. It computes Jaccard similarities over canonical entities ($J_E$), event types ($J_V$), normalized calendar dates ($J_T$), and lexical tokens ($J_L$):

\begin{equation}
\operatorname{sim}(d_i,d_j)=0.35J_E+0.25J_V+0.15J_T+0.25J_L.
\label{eq:similarity}
\end{equation}

An undirected link is created when the score meets the configured alignment threshold. Connected components of this graph become document clusters. This approach is transparent and inexpensive, although its transitive closure can join documents whose pairwise similarity falls below the threshold; this limitation is discussed in Section~\ref{sec:limitations}.

Finally, information fusion computes a cluster title, an extractive summary of the combined texts, the five most frequent entities, the five most frequent event types, and a confidence score. For multi-document clusters, confidence is the mean pairwise similarity expressed as a percentage. Singleton clusters receive a confidence of 100 by construction; this value indicates structural certainty rather than independent corroboration.

\subsection{Layer 5: relational persistence}

SQLite is used as a self-contained central store. Table~\ref{tab:schema} summarizes the schema. Foreign keys connect all semantic annotations to their source documents, while cascade deletion removes dependent rows. Documents and entity edges are updated through idempotent upserts; cluster tables are rebuilt after batch interlinking so that their membership reflects the latest processed set.

\begin{table}[t]
\centering
\caption{Persistent data model.}
\label{tab:schema}
\small
\begin{tabularx}{\columnwidth}{@{}lX@{}}
\toprule
\textbf{Table} & \textbf{Purpose} \\
\midrule
\code{documents} & Source content, metadata, category, sentiment, and summary. \\
\code{entities} & Surface forms, labels, canonical forms, and character spans. \\
\code{temporals} & Temporal mentions, normalized values, and spans. \\
\code{events} & Event types, triggers, sentences, times, and confidence. \\
\code{entity\_edges} & Weighted co-occurrence relationships. \\
\code{clusters} & Fused titles, summaries, confidence, entities, and event types. \\
\code{cluster\_members} & Many-to-many links between clusters and documents. \\
\bottomrule
\end{tabularx}
\end{table}

The database boundary is central to application independence. A future Web service, notebook, batch report, or visualization tool can read the same normalized state without modifying the semantic pipeline.

\section{Grounded Question-Answering Paths}

\system{} extends the monitoring pipeline with three backend question-answering paths. Each returns a structured object containing the question, answer, evidence, model identifier, and scope.

\subsection{Question-driven Web QA}

The Web path augments a question with energy terms when necessary, then attempts multiple search providers. Candidate URLs are scraped into normalized documents. Each page must pass the same energy categorization and semantic extraction used during scheduled ingestion. Accepted pages become bounded evidence items containing a title, source, URL, category, score, summary, entities, events, and excerpt. Evidence may optionally be persisted, thereby allowing a question-driven discovery to enrich the monitored corpus.

\subsection{Database QA}

The database path retrieves either the full stored corpus or an explicit list of document identifiers. It then applies lexical ranking over the title, summary, and text. For query token $q$ and document $d$, the contribution is

\begin{equation}
R(q,d)=5n_t(q,d)+3n_s(q,d)+\mathbb{I}[n_b>0]\left(1+\ln(1+n_b)\right),
\end{equation}

where $n_t$, $n_s$, and $n_b$ denote occurrences in the title, summary, and body. The highest-ranked documents form the evidence set. This transparent baseline is reproducible and avoids introducing an embedding index before its value has been measured.

\subsection{Selected-resource QA}

The selected-resource path accepts indexes of configured sources, fetches only those sources, processes and stores relevant documents, and delegates answering to database QA constrained to the newly stored identifiers. It therefore combines controlled acquisition with the consistent retrieval semantics of the central corpus.

\subsection{Answer synthesis and graceful degradation}

When a language-model API key is configured, evidence blocks are sent to the selected model with an instruction to answer only from the supplied evidence and to cite bracketed source numbers. When no model is available, or when synthesis fails, the backend returns an extractive evidence summary. The fallback is important architecturally: collection, semantic analysis, persistence, retrieval, and inspectable evidence remain available without external model access.

The language model is intentionally placed after retrieval rather than inside the semantic extractors. This limits its authority to presentation and synthesis, preserves deterministic structured annotations, and supports independent evaluation of retrieval quality and answer faithfulness.

\section{Relationship to the Original EneMonIE Design}

The original EneMonIE proposal \citep{kucuk2022} defines the conceptual foundation, while the present architecture introduces the concrete implementation choices summarized in Table~\ref{tab:comparison}. The table distinguishes retained, concretized, and newly introduced elements. The new architecture preserves the semantic intent of EneMonIE while narrowing its boundary to reusable backend services.

\begin{table*}[t]
\centering
\caption{Relationship between the original proposal and the implementation-derived architecture.}
\label{tab:comparison}
\small
\begin{tabularx}{0.97\textwidth}{@{}p{0.19\textwidth}p{0.32\textwidth}X@{}}
\toprule
\textbf{Concern} & \textbf{Original EneMonIE proposal} & \textbf{Implementation-derived architecture} \\
\midrule
Sources & News, videos, social media, scholarly articles, organizational sites, and open data. & Concrete adapters for RSS, URL lists, arXiv, JSONL, and CSV, plus question-driven Web search and scraping. \\
Filtering & Ontology-based text categorization. & Weighted keyword scoring with a configurable relevance threshold before downstream processing. \\
Information extraction & NER, temporal expressions, and energy-event extraction. & Explicit rule- and gazetteer-based modules returning typed objects with spans, canonical values, times, and confidence. \\
Sentiment & Opinion analysis for social-media text. & Source-sensitive English/French lexicon analysis applied only to social-media documents. \\
Interlinking and fusion & Align related media, fuse information, and summarize it. & Weighted Jaccard similarity, threshold graph, connected-component clusters, mean-similarity confidence, and extractive fusion. \\
Knowledge structure & Social-network construction and a central relational database. & Entity co-occurrence edges and a seven-table SQLite schema with foreign keys and cascade deletion. \\
Information access & Server-side application and Web retrieval/visualization. & Three backend QA paths with structured evidence; UI, server, and visualization layers are excluded. \\
Model dependency & Not specified. & Optional language-model synthesis with deterministic extractive fallback. \\
\bottomrule
\end{tabularx}
\end{table*}

The exclusion of the application layer is not a reduction of semantic functionality. It is a boundary decision. The original retrieval and visualization goals can be implemented by one or more downstream consumers, while the backend remains testable without an HTTP server or graphical environment.

\section{Implementation and Verification}

\subsection{Reference implementation}

The prototype is implemented in Python and organized around four main packages: ingestion, processing, question answering, and resources. The \code{EneMonIEPipeline} class composes these packages and owns the central database. Lightweight data classes define documents, entities, temporal expressions, events, sentiment results, processed documents, and clusters. Configuration is loaded from YAML, making sources and thresholds deployment-specific.

\subsection{Preliminary Evaluation: Methodology and Results}
\label{sec:preliminary-evaluation}

An offline evaluation assessed relevance filtering, information extraction, evidence retrieval, and local processing latency. The experiment used a small synthetic fixture to verify implementation behavior; it does not constitute a representative benchmark of real-world energy information extraction.

\paragraph{Dataset and protocol.}
The evaluation comprised seven English energy-relevant documents: six news-style records and one social-media record. Reference annotations covered ten entity mentions, seven sentence-event-type pairs, and five temporal expressions. The question set comprised six questions: five with annotated relevant documents and one deliberately unanswerable question.

The relevance threshold was fixed at 1.5, and temporal interpretation used a reference clock of 22 September 2026 at 00:00:00. Evaluation used a temporary SQLite database, with document creation timestamps fixed according to input order. No network acquisition or language-model synthesis was invoked. Component evaluation applied the extractors directly to the reference-relevant documents, while pipeline evaluation assessed outputs after relevance filtering.

\paragraph{Metrics.}
Precision, recall, and $F_1$ were calculated from pooled true-positive, false-positive, and false-negative counts. Entity matching required the exact character span and entity type; event matching required the document, sentence, and event type; temporal detection required the exact span. Normalized-date accuracy measured calendar-date agreement among correctly matched temporal spans.

Retrieval recall@5 measured the fraction of annotated relevant documents returned within the first five results for each question. MRR@5 measured the reciprocal rank of the first relevant result, assigning zero when none was retrieved. Both metrics were averaged over the five answerable questions. The unanswerable question was assessed separately.

6:19 PM

\begin{table}[tb]
\centering
\caption{Extraction results on seven recovered synthetic documents.}
\label{tab:preliminary-quality}
\small
\setlength{\tabcolsep}{5pt}
\begin{tabular}{@{}lrrrr@{}}
\toprule
\textbf{Task} & $N$ & \textbf{P} & \textbf{R} & $F_1$ \\
\midrule
Entities & 10 & 0.588 & 1.000 & 0.741 \\
Events & 7 & 1.000 & 1.000 & 1.000 \\
Temporal spans & 5 & 1.000 & 1.000 & 1.000 \\
\bottomrule
\end{tabular}
\end{table}

\paragraph{Quality results.}
All seven recovered relevant documents passed the relevance gate, yielding relevance recall of 1.000 on this subset. Because the rerun contains no off-topic documents, it does not establish the filter's ability to reject irrelevant inputs.

Table~\ref{tab:preliminary-quality} summarizes extraction performance. Entity extraction recovered all ten reference mentions, with precision of 0.588 and $F_1$ of 0.741. All seven annotated event pairs and all five temporal spans were recovered without additional predictions in those tasks. Normalized-date accuracy was $5/5=1.000$. Relative expressions, including ``Monday'' and ``yesterday,'' were normalized to 21 September 2026 under the reference clock.

On the recovered synthetic cases, the pipeline achieved an entity $F_1$ of 0.741, an event $F_1$ of 1.000, and a temporal-span $F_1$ of 1.000.

Recall@5 and MRR@5 both reached 1.000. Every answerable question retrieved all annotated relevant documents within the first five results, with a relevant document ranked first. The deliberately unanswerable question returned no evidence. These outcomes demonstrate retrieval behavior on the supplied questions, without establishing general answer correctness or reliable abstention across other queries.

\paragraph{Runtime.}
On Windows 11 with Python 3.13.14, one warmup call per document preceded five measured processing calls per document, producing 35 timing observations. Median processing latency was approximately 0.183,ms. These measurements exclude initialization, warmup, storage, retrieval, and language-model calls. They concern very short synthetic documents and do not estimate application throughput.

\paragraph{Interpretation and reproducibility.}
The evaluation script and its JSON and Markdown reports record configuration, predictions, extraction counts, query rankings, runtime scope, and reference provenance. The results support the intended behavior of the event, temporal, and retrieval components on the evaluated fixtures. The small synthetic corpus and reconstructed annotations limit broader conclusions. Evaluation on independently annotated real documents with development/test separation remains necessary. Sentiment quality, interlinking, network reliability, and generated-answer quality were outside the scope of this experiment.

4:38 PM

\section{Discussion}

\subsection{Architectural benefits}

The redesign provides three practical benefits. First, it establishes a single semantic path for scheduled monitoring and question-driven discovery. As a result, evidence collected for a question is filtered and represented consistently with the rest of the corpus. Second, it separates deterministic semantic state from optional generative synthesis. Database rows, event confidence, entity canonicalization, and cluster membership can be inspected independently of a model-generated answer. Third, it gives downstream applications a stable relational and object-level contract instead of coupling them to source formats.

The architecture is also suitable for incremental modernization. Rule-based NER can be replaced by a transformer model; lexical ranking can be replaced by hybrid sparse-dense retrieval; SQLite can be migrated to a server database; and connected-component clustering can be replaced by a time-aware or learned method. Because each responsibility is localized, these changes need not alter all consumers simultaneously.

\subsection{Limitations}
\label{sec:limitations}

The current system remains a research prototype and therefore presents several limitations. First, it relies heavily on lexical methods. Weighted keywords, gazetteers, and trigger lists are interpretable, but they provide limited recall for paraphrases, spelling variations, and emerging terminology. Additionally, the temporal component supports only a bounded set of explicit date patterns and does not resolve complex relative expressions or event-specific temporal scopes.

Another limitation concerns sentiment analysis. Lexicon-based counts do not adequately capture negation, sarcasm, opinion targets, or aspect-specific sentiment. The system also relies on transitive clustering, which may cause connected components to group chains of documents even when the first and last documents are only weakly related. Moreover, confidence assigned to singleton clusters should not be interpreted as evidence of corroboration.

\subsection{Evaluation agenda}

Future work should evaluate each stage separately and end to end. Relevance filtering can be measured with precision, recall, and $F_1$ on a stratified energy/non-energy corpus. Entity, temporal, and event extraction require span- and type-level annotations. Interlinking should be evaluated using document-pair and cluster metrics, with attention to time-separated events sharing the same entities. Retrieval can be measured with recall@$k$, mean reciprocal rank, or normalized discounted cumulative gain. Grounded answer quality should include factual consistency, citation correctness, evidence sufficiency, and abstention when the corpus is incomplete.

Scalability experiments should report ingestion throughput, per-document processing time, database growth, pairwise clustering cost, and the effect of source failures. Because the current interlinker compares all document pairs, approximate candidate generation or blocking by time, entity, and category will become necessary as the corpus grows.

Finally, multilingual expansion should move beyond isolated English and French lexicons. Language identification, multilingual entity normalization, cross-lingual clustering, and evaluation on comparable reports are necessary to fulfill the original multilingual ambition of EneMonIE.

\section{Conclusion}

This article introduced \system, an executable backend reinterpretation of the EneMonIE energy-monitoring architecture. The design retains the original semantic objectives while making acquisition, normalization, relevance filtering, extraction, interlinking, fusion, persistence, and question answering operational. Its five-layer decomposition is complemented by three evidence-grounded QA paths that reuse the same semantic pipeline. The deliberate exclusion of user-interface and server-side application components creates a reusable core that can support multiple future applications.

The implementation demonstrates architectural feasibility and passes a deterministic integration smoke test. It does not yet establish large-scale extraction or question-answering performance. The next scientific step is therefore a component-wise and end-to-end evaluation on labeled, multilingual energy corpora, followed by targeted replacement of rule-based modules where measured limitations justify greater model complexity.

\section*{Reproducibility Statement}

The accompanying Overleaf project contains this manuscript, the architecture figure, and the BibTeX database. The described validation is deterministic and does not require an external language model. Live Web acquisition remains dependent on third-party source availability and should be evaluated with recorded URLs and retrieval timestamps.

\balance
\bibliographystyle{plainnat}
\bibliography{references}

\end{document}